\documentclass{article}

\usepackage{PRIMEarxiv}

\usepackage[utf8]{inputenc} 
\usepackage[T1]{fontenc}    
\usepackage{hyperref}       
\usepackage{url}            
\usepackage{booktabs}       
\usepackage{amsfonts}       
\usepackage{nicefrac}       
\usepackage{microtype}      
\usepackage{lipsum}
\usepackage{fancyhdr}       
\usepackage{graphicx}       
\graphicspath{{media/}}     

\usepackage{todonotes}
\newcommand{\modelname}{Albertina}
\newcommand{\deberta}{DeBERTa}
\newcommand{\bertimbau}{BERTimbau}
\newcommand{\brwac}{brWaC}
\newcommand{\ptpt}{\mbox{PT-PT}}
\newcommand{\ptbr}{\mbox{PT-BR}}

\title{Advancing Neural Encoding of Portuguese\\with Transformer \modelname\ PT-*
}
\author{João Rodrigues,$^{\diamondsuit}$ Luís Gomes,$^{\diamondsuit}$ João Silva,$^{\diamondsuit}$ António Branco,$^{\diamondsuit}$ \\
{\bf Rodrigo Santos},$^{\diamondsuit}$ {\bf Henrique Lopes~Cardoso},$^{\heartsuit}$ {\bf Tomás Osório}$^{\heartsuit}$   \\
   $^{\diamondsuit}$University of Lisbon\\
NLX – Natural Language and Speech Group, Dept of Informatics\\
Faculdade de Ciências (FCUL), Campo Grande, 1749-016 Lisboa, Portugal \\
  $^{\heartsuit}$Laboratório de Inteligência Artificial e Ciência de Computadores (LIACC)\\
  Faculdade de Engenharia da Universidade do Porto (FEUP)\\Rua Dr. Roberto Frias, 4200-465 Porto, Portugal \\
  }

\begin{document}
\maketitle

\begin{abstract}
To advance the neural encoding of Portuguese (PT), and a fortiori the technological preparation of this language for the digital age, we developed a Transformer-based foundation model that sets a new state of the art in this respect for two of its variants, namely European Portuguese from Portugal (\ptpt) and American Portuguese from Brazil (\ptbr). 

To develop this encoder, which we named \modelname\ PT-*, a strong model was used as a starting point, \deberta, and its pre-training was done over data sets of Portuguese, namely over data sets we gathered for \ptpt\ and for \ptbr\, and over the \brwac\ corpus for \ptbr.
The performance of \modelname\ and competing models was assessed by evaluating them on prominent downstream language processing tasks adapted for Portuguese.

Both \modelname\ \ptpt\ and \ptbr\ versions are distributed free of charge and under the most permissive license possible and can be run on consumer-grade hardware, thus seeking to contribute to the advancement of research and innovation in
language technology for Portuguese.

\end{abstract}


\section{Introduction}
\label{sec:introduction}

In recent years, the field of Artificial Intelligence has come to successfully exploit the paradigm of deep learning, a machine learning approach based on large artificial neural networks~\cite{LeCun:2015:dl}.
Applied to Natural Language Processing (NLP), deep learning gained outstanding traction with notable breakthroughs under the distributional semantics approach, namely with word embedding techniques~\cite{Mikolov:2013:embeddings} and the Transformer neural architecture~\cite{Vaswani:2017:attention}.
These neural models acquire semantic representations from massive amounts of data in a self-supervised learning process that ultimately results in the so-called \emph{Foundation Models}~\cite{Bommasani:2021:opportunitiesandrisks}.

Self-supervision is accomplished in NLP through language modeling~\cite{Bengio:2000:neuralmodel} and was initially adopted in shallow neural network models such as Word2Vec~\cite{Mikolov:2013:embeddings} for the creation of word embeddings.
Over time, this approach was scaled beyond the single-token level to sequence transduction with encoding-decoding models based on recurrent~\cite{Sutskever:2014:seq2seq} or convolution neural networks and occasionally supported by attention mechanisms~\cite{Bahdanau:2014:nmtattention}.

A particular neural network architecture, the Transformer, has stood out among all others, showing superior performance by a large margin, sometimes even surpassing human-level performance~\cite{Wang:2018:glue,Wang:2019:superglue}, and became mainstream in virtually every NLP task and application~\cite{Bommasani:2021:opportunitiesandrisks}.
Several variants have spun out from the base Transformer architecture (encoder-decoder), including the landmark encoder BERT~\cite{Devlin:2018:bert} and the outstanding decoder GPT~\cite{Brown:2020:fewshotlearners}, which have been most successfully adapted to downstream tasks, complemented with techniques such as transfer learning~\cite{pan2010survey}, fine-tuning~\cite{Peters:2019:tune} or few-shot prompting~\cite{Brown:2020:fewshotlearners}.

The large scale of foundation models is crucial to their emergent capabilities and successful deployment.

Adding to the difficulty of accessing sufficiently large and powerful computational resources, most NLP research is focused on the English language, which is just one of the around 7,000 idioms on the planet.
Consequently, there is a lack of competitive and openly available foundation models specifically developed for the vast majority of languages other than English, which happens to be also the case for the Portuguese language.
This restrains the scientific progress and the innovative exploitation related to those languages, as well as curtailing other societal impacts and benefits, further enlarging the digital language divide between English and other languages.

To the best of our knowledge, there are a couple of publicly published models that were developed specifically for Portuguese, namely for its European variant from Portugal (\ptpt) and its American variant from Brazil (\ptbr).
However, they present considerable drawbacks,
namely in what concerns their sub-optimal performance level
and the non-existent public distribution of encoders for the \ptpt\ variant.

Accordingly, there is important motivation and considerable room for improvement in creating new and better encoders for Portuguese, which we developed and present in this paper --- and named as \modelname\ PT-*.\footnote{The Albertina-PT-PT model can be obtained here: \url{https://huggingface.co/PORTULAN/albertina-ptpt} and the Albertina-PT-BR model can be obtained here: \url{https://huggingface.co/PORTULAN/albertina-ptbr}} %
On a par with an encoder for \ptbr\ that sets a new state of the art for this language variant, its twin \ptpt\ version is an original contribution to the state-of-the-art concerning Portuguese: a freely available neural encoder specifically developed for its European variant with highly competitive performance, whose reporting publication is publicly available and which is openly distributed.

The remainder of this paper is organized as follows. Section~\ref{sec:relatedwork} provides an overview of existing models with support for Portuguese, with a particular focus on the pre-existing \bertimbau, for \ptbr.
The data sets used in pre-training and evaluating our model are presented in Section~\ref{sec:datasets}.
Section~\ref{sec:foundationmodel} describes \modelname\ PT-* and its pre-training and fine-tuning procedures.
The evaluation results of its versions on downstream tasks are discussed in Section~\ref{sec:results}.
Section~\ref{sec:finalremarks} closes the paper with concluding remarks.

\section{Related Work}
\label{sec:relatedwork}

Regarding related work, we consider Transformer-based encoder models that, to the best of our knowledge, are concerned with the Portuguese language.
Accordingly, besides searching the literature, we also screened the Hugging Face~\cite{HuggingFace:website} model repository, as it has become the main source of NLP models.

Multiple studies~\cite{Devlin:2018:bert,Virtanen:2019:bertfinnish,DeVries:2019:bertje,Martin:2019:camembert,Souza:2020:bertimbau,Armengol:2021:multilingualmodels} have shown that language-specific foundation models perform better than multilingual ones.
This realization has thus led to a few initiatives that created language-specific encoders, trained from scratch for a single language, such as
BERTa for Catalan~\cite{Armengol:2021:multilingualmodels},
ERNIE for Chinese~\cite{Sun:2021:ernie3},
BERTje for Dutch~\cite{DeVries:2019:bertje},
FinBERT for Finnish~\cite{Virtanen:2019:bertfinnish},
CamemBERT for French~\cite{Martin:2019:camembert},
and
MarIA for Spanish~\cite{Gutierrez:2022:maria}, among others.

Nevertheless, given it is not always viable to create a model specifically for a given language due to a lack of available data or computing resources, multilingual models have been resorted to as a temporary yet common mitigation for this problem for many languages.
These are models that are pre-trained on data that include a mix of languages---albeit English is typically present in a greater amount---and are thus capable of modeling multiple languages.

\subsection{Encoders whose multilingual data set included Portuguese}
\label{sec:relatedwork:monovsmulti}

Taking the number of Hugging Face downloads as a proxy for popularity and user base size, the stand-out models that support Portuguese are multilingual, namely XML-Roberta, available in base and large sizes, Multilingual BERT (mBERT) base cased, and DistilBERT base.

Several task-specific or domain-specific models have been built upon these multilingual foundations.
For instance, BioBERTpt (Portuguese Clinical and Biomedical BERT)~\cite{Schneider:2020:biobertpt} was created by fine-tuning mBERT on clinical notes and biomedical literature in Portuguese.

\subsection{Encoders specifically concerned with Portuguese}
\label{sec:relatedwork:portuguese}

To the best of our knowledge, for \ptpt\ there is the encoder presented in \cite{miquelina2022generating}, but it is not possible to find therein clear evaluation results against prominent downstream tasks and, most importantly, the distribution of that model is not announced.

As for \ptbr, there are a couple of encoders publicly distributed. That is the case of BERTabaporu,\footnote{\url{https://huggingface.co/pablocosta/bertabaporu-base-uncased}} which is of limited interest though, given its quite narrow domain, as it is a BERT-based encoder trained on Twitter data.
The most popular of these two encoder models for \ptbr, by far, is \bertimbau~\cite{Souza:2020:bertimbau}.

\bertimbau\ is available in two model sizes, base, with 110~million parameters, and large, with 330~million parameters.
In both cases, the authors took an existing BERT-based model as starting point and, after discarding the word embeddings and the masked language modeling head layers, performed a hefty 1~million steps of additional pre-training on the \brwac\ corpus (see Section~\ref{sec:datasets:pretraining}).

\begin{itemize}
    \item \bertimbau\ base took multilingual mBERT base~\cite{Devlin:2018:bert} as its starting point. It was pre-trained with a batch size of 128 and sequences of 512 tokens during 4~days on a TPU v3-8 instance, performing about 8~epochs on the corpus~\cite[\S 5.1]{Souza:2020:bertimbau}.
    \item \bertimbau\ large took the monolingual English BERT large~\cite{Devlin:2018:bert} as the starting point, given there was no multilingual mBERT available in large size. It was pre-trained with sequences of 128 tokens in batches of size 256 for the first 900,000 steps and sequences of 512 tokens in batches of size 128 for the final 100,000 steps. Its pre-training took 7~days on a TPU v3-8 instance and performed about 6~epochs on the corpus~\cite[\S 5.1]{Souza:2020:bertimbau}.
\end{itemize}

Both the base and large variants of \bertimbau\ outperform mBERT in a couple of downstream tasks in Portuguese, with the large variant being better~\cite{Souza:2020:bertimbau}.
Given this was an inaugural general-domain encoder for Portuguese, it set the state of the art for those tasks in Portuguese.\footnote{As such, \bertimbau\ has come to serve as the basis for several other task-specific models available in Hugging Face.
These task-specific models, however, appear to be unpublished, unnamed, or provide no information on their Hugging Face page; as such, they will not be covered in the present paper.}

Since the creation of \bertimbau, improved Transformer-based architectures have been developed, which, together with more efficient training techniques, should allow better-performing models to be developed.
This strengthens the motivation to develop and distribute alternative, state-of-the-art encoders also for \ptbr.

\section{Data sets}
\label{sec:datasets}

We proceed now with presenting the data sets used to pre-train \modelname~PT-* and the data sets used to fine-tune it for the downstream tasks where it was extrinsically evaluated, for both \ptpt\ and \ptbr\ variants.

\subsection{Data sets for the pre-training stage}
\label{sec:datasets:pretraining}

To secure conditions for comparability with \bertimbau, for the pre-training of the \modelname\ \ptbr\ we resorted to the same data set, the \brwac\ corpus (Brazilian Portuguese Web as Corpus)~\cite{Wagner:2018:brwac}.
This corpus contains 2.7~billion tokens in 3.5~million documents and was obtained from crawling many different sites to ensure diversity.
The authors report that some effort was made to remove duplicated content.

As for the pre-training of the \modelname\ \ptpt, we resorted to a data set that resulted from gathering some openly available corpora of European Portuguese from the following sources:
\begin{itemize}
    \item OSCAR~\cite{abadji-etal-2022-towards}: the OSCAR data set includes documents in more than one hundred languages, including Portuguese, and it is widely used in the literature. It is the result of a selection performed over the Common Crawl\footnote{\url{https://commoncrawl.org/}} data set, crawled from the Web, that retains only pages whose metadata indicates permission to be crawled, that performs deduplication, and that removes some boilerplate, among other filters. Given that it does not discriminate between the Portuguese variants, we performed extra filtering by retaining only documents whose meta-data indicate the Internet country code top-level domain of Portugal. We used the January 2023 version of OSCAR, which is based on the November/December 2022 version of Common Crawl.
    \item DCEP~\cite{hajlaoui-etal-2014-dcep}: the Digital Corpus of the European Parliament is a multilingual corpus including documents in all official EU languages published on the European Parliament's official website. We retained its European Portuguese portion.
    \item Europarl~\cite{koehn-2005-europarl}: the European Parliament Proceedings Parallel Corpus is extracted from the proceedings of the European Parliament from 1996 to 2011. We retained its European Portuguese portion.
    \item ParlamentoPT: the ParlamentoPT is a data set we obtained by gathering the publicly available documents with the transcription of the debates in the Portuguese Parliament.
\end{itemize}

We filtered these data using the BLOOM~\cite{bigscience-roots:2022} pre-processing pipeline,\footnote{We skipped the default filtering of stopwords since it would disrupt the syntactic structure, and also the filtering for language identification given the corpus was pre-selected as Portuguese.} resulting in a data set of 8~million documents, containing around 2.2~billion tokens.
The number of documents from each source---Europarl, DCEP, ParlamentoPT, and OSCAR data---corresponds approximately to 15\%, 20\%, 31\%, and 34\% of the entire data set for \ptpt, respectively. All these data sets are publicly available, including ParlamentoPT.\footnote{ParlamentoPT was collected from the Portuguese Parliament portal in accordance with its open data policy (\url{https://www.parlamento.pt/Cidadania/Paginas/DadosAbertos.aspx}, and can be obtained here: \url{https://huggingface.co/datasets/PORTULAN/parlamento-pt}.}

\subsection{Data sets for the fine-tuning concerning downstream tasks}
\label{sec:datasets:downstream}

We organized the data sets used for downstream tasks into two groups.
In one group, we have the two data sets from the ASSIN~2 benchmark, namely STS and RTE, that were used to evaluate \bertimbau~\cite{Souza:2020:bertimbau}.

In the other group of data sets, we have the translations into \ptbr\ and \ptpt\ of the English data sets used for a few of the tasks in the widely-used GLUE benchmark~\cite{Wang:2018:glue}, which allowed us to test both \modelname\ variants on a wider variety of downstream tasks.

\subsection*{ASSIN~2}

ASSIN~2~\cite{Real:2020:assin2} is a \ptbr\ data set of approximately 10,000 sentence pairs, split into 6,500 for training, 500 for validation, and 2,448 for testing, annotated with semantic relatedness scores (range 1 to 5) and with binary entailment judgments.
This data set supports the task of semantic text similarity (STS), which consists of assigning a score of how semantically related two sentences are, and the task of recognizing textual entailment (RTE), which, given a pair of sentences, consists of determining whether the first entails the second.

We did not create a \ptpt\ version of ASSIN~2.
That would require transposing the data set, which is \ptbr, into \ptpt; however, to our knowledge, there is no automatic translation system for direct translation between those variants.
One solution would be to translate through an intermediate language, say English or Spanish, and then translate the result into \ptpt, but doing this would likely highly degrade the quality of the resulting benchmark by a factor that would not be possible to determine.

\subsection*{GLUE tasks translated}

GLUE~\cite{Wang:2018:glue} has become a standard benchmark for model evaluation on downstream tasks.
As the original GLUE is in English, we resort to PLUE~\cite{Gomes:2020:plue} (Portuguese Language Understanding Evaluation), a data set that was obtained by automatically translating GLUE~\cite{Wang:2018:glue} into \ptbr.
We address four tasks from those in PLUE, namely:
\begin{itemize}
    \item two similarity tasks: MRPC, for detecting whether two sentences are paraphrases of each other, and STS-B, for semantic textual similarity;
    \item and two inference tasks: RTE, for recognizing textual entailment,\footnote{This is the same task as the ASSIN~2 RTE, but on different source data.} and WNLI, for coreference and natural language inference.
\end{itemize}

To obtain the \ptpt\ version of this benchmark, we automatically translated the same four tasks from GLUE using DeepL Translate,\footnote{\url{https://www.deepl.com/}} which specifically provides translation from English to \ptpt\ as an option.\footnote{
This benchmark is freely distributed here: \url{https://huggingface.co/datasets/PORTULAN/glue-ptpt}}

\section{\modelname\ PT-* model}
\label{sec:foundationmodel}

We describe the pre-training of the \modelname\ language model for Portuguese, in its two \ptpt\ and \ptbr\ versions, as a continuation of the pre-training of \deberta\ with our data sets.
We also address its fine-tuning for the downstream tasks considered for its extrinsic evaluation.

\subsection{The starting encoder}

We take \deberta~\cite{He:2021:deberta} as our starting encoder since it is reported to improve on multiple strong encoders and surpass human performance on the SuperGLUE benchmark.
The main novelty in \deberta\ comes from two techniques, namely \emph{disentangled attention} and \emph{enhanced mask decoder}, which are related to how information about the relative and the absolute positions of tokens is encoded and handled by the model.

In other BERT-like encoders and Transformers in general, information about the position of tokens is represented as a vector, such as, for instance, a sinusoidal embedding, that is added to the content embedding of the token.
The disentangled attention mechanism in \deberta\ uses separate content (\(H\)) and relative position (\(P\)) embeddings, and the attention mechanism attends separately to these embeddings.
So, when calculating the cross attention \(A_{i,j}\) between tokens \(i\) and \(j\), the disentangled attention mechanism incorporates not only the usual content-to-content attention \(H_iH_j^T\) but also content-to-position \(H_iP_{j|i}^T\) attention and position-to-content \(P_{i|j}H_j^T\) attention.

The second specific mechanism in \deberta, the enhanced mask decoder, incorporates information about the absolute position of tokens right before the softmax layer to predict the masked tokens.
Usually, all three inputs (Query, Key, and Value) to the self-attention calculation come from the hidden states in the preceding layer, but in the enhanced mask decoder of \deberta\ the Query input is based on the absolute position of the token.

As codebase, we resorted to the \deberta\ V2 XLarge, for English, that is available from Hugging Face.\footnote{\url{https://huggingface.co/microsoft/deberta-v2-xlarge}}
We use the Transformers \cite{wolf-etal-2020-transformers} library with accelerate \cite{accelerate}.
It has 24 layers with a hidden size of 1536 and a total of 900~million parameters.
This version brings some changes to the original \deberta\ paper~\cite{He:2021:deberta}. In particular:
(i)~it uses a vocabulary size of 128,000 and the \textit{sentencepiece} tokenizer \cite{kudo-richardson-2018-sentencepiece},
(ii)~it adds an additional convolution layer to the first Transformer layer,
and (iii)~it shares the position projection and content projection matrices in the attention layer.

\subsection{Pre-training \modelname\ \ptbr}

\begin{figure}[h]
    \centering
    \includegraphics[width=0.6\textwidth]{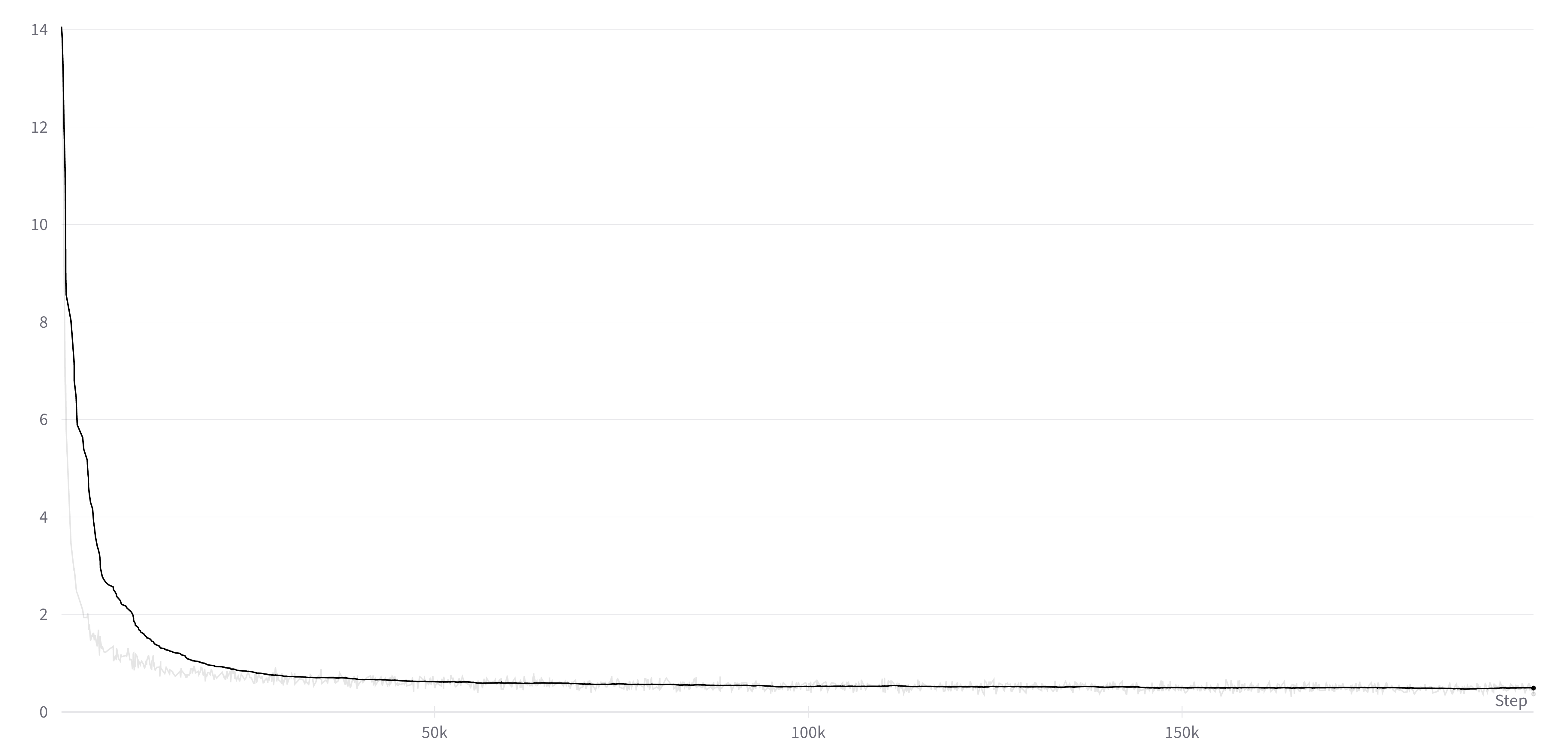}
    \caption{Training loss for \modelname\ \ptbr\, with a smoothing factor of .95 over the exponential moving average.}
    \label{fig:deberta_loss_ptbr}
\end{figure}

For the training of \modelname\ \ptbr, the \brwac\ data set was tokenized with the original \deberta\ tokenizer with a 128-token sequence truncation and dynamic padding.
The model was trained using the maximum available memory capacity\footnote{The \ptbr\ model was trained for 1 day and 11 hours on a2-megagpu-16gb Google Cloud A2 VMs with 16 GPUs, 96 vCPUs and 1.360 GB of RAM.} resulting in a batch size of 896 samples (56 samples per GPU without gradient accumulation steps).
We chose a learning rate of 1e-5 with linear decay and 10k warm-up steps based on the results of exploratory experiments. 
In total, around 200k training steps were taken across 50 epochs.
Additionally, we used the standard BERT masking procedure with a 15\% masking probability for each example. Figure~\ref{fig:deberta_loss_ptbr} illustrates the model's convergence during training.

In order to provide a more permissive licensed model for the PT-BR variant, we also developed the Albertina PT-BR No-brWaC model.\footnote{The Albertina-PT-BR No-brWaC model can be obtained here: \url{https://huggingface.co/PORTULAN/albertina-ptbr-nobrwac}.} 
This model was trained using a curated selection of documents from the OSCAR data set, specifically filtered by the Internet country code top-level domain of Brazil. It adheres to the same filtering pipeline employed in the aforementioned Albertina PT-PT model (Section \ref{sec:datasets:pretraining}).
The resulting data set contains approximately 3.7 billion tokens.
We resorted to the same hyperparameters as the Albertina PT-BR model.

\subsection{Pre-training \modelname\ \ptpt}

\begin{figure}[h]
    \centering
    \includegraphics[width=0.6\textwidth]{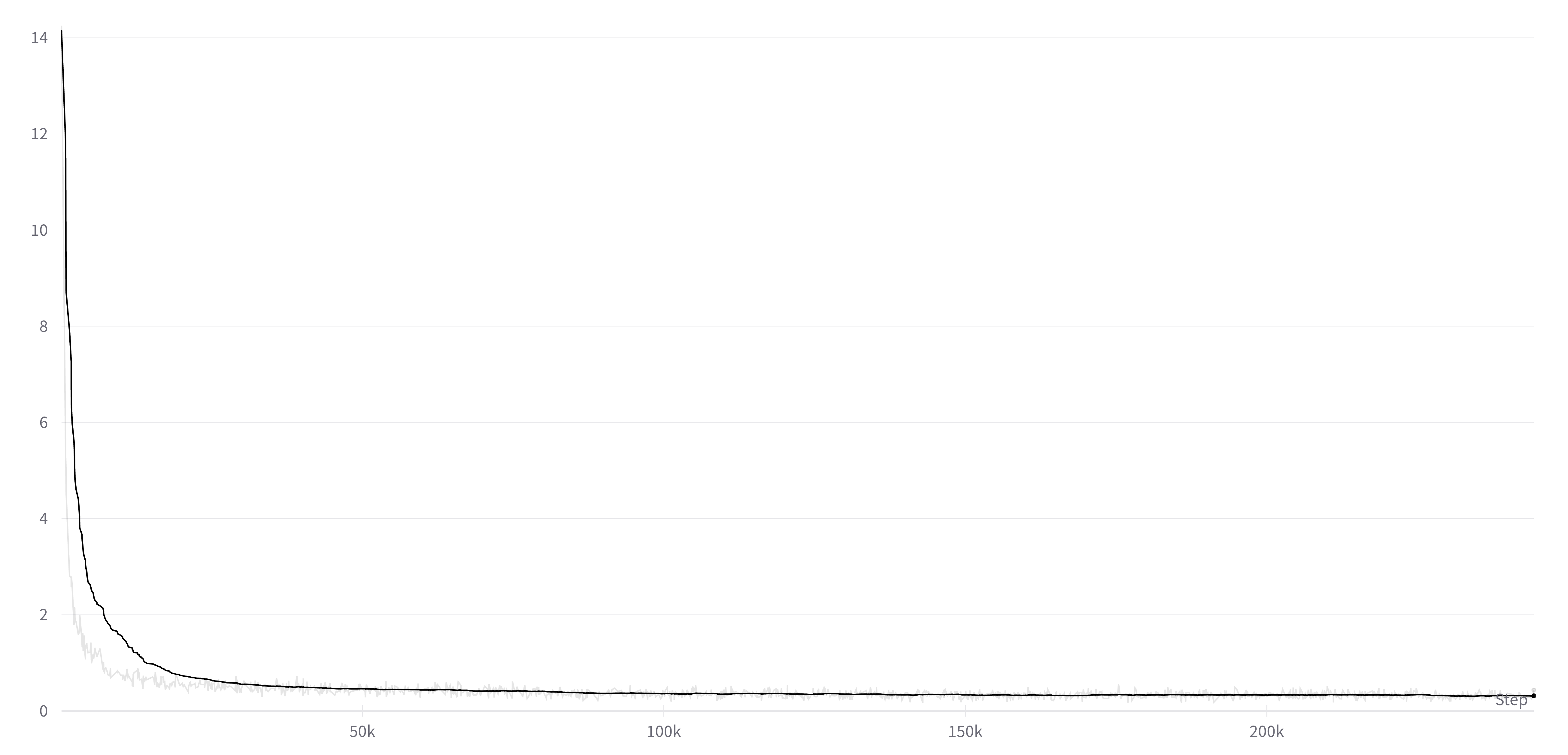}
    \caption{Training loss for \modelname\ \ptpt\, with a smoothing factor of .95 over the exponential moving average.}
    \label{fig:deberta_loss_ptpt}
\end{figure}

To train \modelname\ \ptpt, the data set was tokenized with the original DeBERTa tokenizer. The sequences were truncated to 128 tokens and dynamic padding was used during the training.
The model was trained using the maximum available memory capacity\footnote{The \ptpt\ model was trained for 3 days on a2-highgpu-8gb Google Cloud A2 VMs with 8 GPUs, 96 vCPUs and 680 GB of RAM.} resulting in a batch size of 832 samples (52 samples per GPU and applying gradient accumulation in order to approximate the batch size of the \ptbr\ model).
Similarly to the \ptbr\ variant above, we opted for a learning rate of 1e-5 with linear decay and 10k warm-up steps.
However, since the number of training examples is approximately twice of that in the \ptbr\ variant, we reduced the number of training epochs to half and completed only 25 epochs, which resulted in approximately 245k steps. Figure~\ref{fig:deberta_loss_ptpt} illustrates the model's convergence during training.

\subsection{Pre-training \modelname\ base models}

Additionally, for the sake of convenience, we have developed also two smaller models specifically designed for each variant, Albertina PT-PT base and PT-BR base. These models are built upon the DeBERTa V1 base model, which consists of 100 million parameters. The Albertina PT-PT base model resorts to the same pre-training data as the aforementioned Albertina PT-PT model, while the Albertina PT-BR base model is trained using the same pre-training data as the Albertina PT-BR No-brWaC model.

Both models were trained using the maximum available memory capacity\footnote{Each model was trained for approximately one day on a2-megagpu-16gb Google Cloud A2 VMs with 16 GPUs, 96 vCPUs and 1.360 GB of RAM.} resulting in a batch size of 3072 samples (192 samples per GPU).

For both base models, the data sets were tokenized with the original DeBERTa tokenizer with a 128 token sequence truncation and dynamic padding. The models were trained using the maximum available memory capacity resulting in a batch size of 3072 samples (192 samples per GPU). We opted for a learning rate of 1e-5 with linear decay and 10k warm-up steps. 
The PT-PT model was trained with a total of 200 training epochs, and PT-BR with 150, resulting in approximately 180k steps in both cases.
Each model was trained for one day on a2-megagpu-16gb Google Cloud A2 VMs with 16 GPUs, 96 vCPUs and 1.360 GB of RAM.

\subsection{Fine-tuning \modelname\ and \bertimbau}
\label{sec:finetuning}
The models for \ptbr\ (\modelname\ \ptbr, \modelname\ \ptbr\ No-brWaC, \modelname\ \ptbr\ base and \bertimbau\ large) were fine-tuned for each of the 6 tasks described above (4 from GLUE and 2 from ASSIN~2), while the models for \ptpt\ (\modelname\ \ptpt\ and \modelname\ \ptpt\ base) were fine-tuned on the 4 GLUE tasks only (as ASSIN-2 is for \ptbr).
Each of these model-task combinations was fine-tuned for a range of sets of hyper-parameter values, with the purpose of selecting the best-performing set of hyper-parameters for each combination.
Specifically, we experimented with dropout 0 and 0.1, learning rate 1e-6, 5e-6 and 1e-5, 32bit and 16bit floating point precision, and random seeds 41, 42, and 43.  When combined, these ranges resulted in a considerable experimental space, with 36 experiments for each model-task pair.
In every such experiment, the whole model was fine-tuned (not just its output head), for 5 epochs with batches of 16 examples.

\section{Experimental Results}
\label{sec:results}

The experimental results obtained are reported in this section. Every score reported is the average of three runs with different seeds. The set of hyper-parameters that produced the highest score on the development data for a given model/task was selected to subsequently evaluate it. It is the corresponding score over the test data that is reported.

\subsection{Improving the state of the art on ASSIN~2 tasks}
\label{sec:results:assin}

The performance scores of the models for \ptbr\ on the RTE task and STS task of ASSIN~2 are displayed in Table~\ref{tab:assin2}.  Our model improves the state of the art for \ptbr\ on these two tasks by a quite competitive margin. 

\begin{table}[h]
    \setlength{\tabcolsep}{1.8ex}
    \centering
    
    \begin{tabular}{lcc}
        \toprule          & RTE             & STS             \\ \midrule
        \modelname\ \ptbr & \textbf{0.9130} & \textbf{0.8676} \\
                Albertina PT-BR No-brWaC & 0.8950 & 0.8547 \\
        \bertimbau\ large &         0.8913  &         0.8531  \\

                \midrule
        Albertina PT-BR base & 0.8652 & 0.8305 \\
    \end{tabular}
    \vspace{0.3cm}
    \caption{Performance on the ASSIN~2 tasks RTE (Accuracy) and STS (Pearson). Higher values indicate better performance, with the best results in bold.}
    \label{tab:assin2}
\end{table}

\subsection{Setting the state of the art on Portuguese GLUE tasks}
\label{sec:results:ptbr}

The performance of \modelname\ and \bertimbau\ large are compared again, this time on the four tasks from PLUE, in \ptbr. As displayed in Table~\ref{tab:tasksptbr}, our Albertina PT-BR model continues to show superior performance, in three of these four tasks.

\begin{table*}[h]
    \setlength{\tabcolsep}{1.8ex}
    \centering
    
    \begin{tabular}{lcccc}
        \toprule          &            RTE  &           WNLI  &           MRPC  &          STS-B  \\ \midrule
                 Albertina PT-BR No-brWaC & 0.7798 & 0.5070 & \textbf{0.9167} & 0.8743 \\
        \modelname\ \ptbr &         0.7545  &         0.4601  &         0.9071  & \textbf{0.8910} \\
        \bertimbau\ large &         0.6546  & \textbf{0.5634} &         0.8873  &         0.8842  \\ \midrule
                 Albertina PT-BR base & 0.6462 & 0.5493 & 0.8779 & 0.8501 \\ 
   \bottomrule

     \modelname\ \ptpt & \textbf{0.7960} &         0.4507  & 0.9151 &         0.8799  \\ 
         Albertina PT-PT base & 0.6643 & 0.4366 & 0.8966 & 0.8608 \\
        
    \end{tabular}
    \caption{Performance on the PLUE tasks, for \ptbr, namely RTE and WNLI (Accuracy), MRPC (F1) and STS-B (Pearson).}
    \label{tab:tasksptbr}
\end{table*}

Table~\ref{tab:tasksptpt} shows the performance of \modelname\ on the same four tasks from GLUE as before, but now automatically translated to \ptpt.

\begin{table*}[h]
    \setlength{\tabcolsep}{1.8ex}
    \centering
    
    \begin{tabular}{lcccc}
        \toprule          &            RTE  &           WNLI  &           MRPC  &          STS-B  \\ \midrule
        \modelname\ \ptpt & \textbf{0.8339} & 0.4225 & \textbf{0.9171} &         0.8801  \\ \midrule
                Albertina PT-PT base & 0.6787 & 0.4507 & 0.8829 & 0.8581 \\
         \bottomrule
\modelname\ \ptbr &         0.7942  &         0.4085  &         0.9048  & \textbf{0.8847} \\
        Albertina PT-BR base & 0.6570 & \textbf{0.5070} & 0.8900 & 0.8516 \\ 
    \end{tabular}
    \caption{Performance on the GLUE tasks translated into \ptpt, namely RTE and WNLI (Accuracy), MRPC (F1) and STS-B (Pearson).}
    \label{tab:tasksptpt}
\end{table*}

\subsection{Discussion}

In this study, we present a Transformer-based foundation model that establishes a new state-of-the-art performance for multiple benchmark data sets in Portuguese. 
It is worth noting that the better efficacy of our model, compared to the pre-existing \bertimbau, goes on par with its better efficiency, as efficacy is achieved with significantly reduced computational requirements compared to pre-existing models. In particular, while the \bertimbau\ model was trained over one million steps, our model required less than a quarter of a million steps.
Our model's ability to achieve superior performance with less training time/computation likely results from resorting to all pre-trained layers, including the first layer, concerning word embeddings, and the last layer, concerning masked token prediction (the masked language modeling head), in contrast to the common practice in the literature of resetting these two layers to random weights to continue the pre-training.

With the cross-evaluation, the motivation for the creation of separated versions for the two language variants \ptpt\ and \ptbr\ is somewhat empirically justified:
when evaluated on \ptpt\ tasks, \modelname\ \ptpt\ outperforms \modelname\ \ptbr\ in all tasks except one, where it is only marginally inferior, cf.~Table~\ref{tab:tasksptpt}; conversely,
when evaluated on \ptbr\ data, \modelname\ \ptbr\ outperforms \modelname\ \ptpt\ in half of the tasks, and \modelname\ \ptbr\ No-brWac in another one, cf.~Table~\ref{tab:tasksptbr}.

As expected given its smaller dimension, the 100M base models are outperform by the 900M parameter models.

Although not directly comparable, the state-of-the-art English models using the original GLUE data sets\footnote{\url{https://gluebenchmark.com/leaderboard}} show performance results that are slightly superior to the results with \modelname.
We hypothesized that this is due mainly to the fact that the English models were evaluated on the respective GLUE test sets (by being submitted to the automatic GLUE benchmark online), while \modelname\ was not.
The reason was that the GLUE online service for testing was not available when we needed it and provided no notice about whether it would reopen. We had thus to evaluate our model offline, and thus on a different split of the data. We used the original development set for evaluation, and from the original training set, we used 10\% for development and the rest for actual training.
Moreover, we consider that the WNLI task was particularly affected by this difference in data partition given its limited sample size (being the smallest of the data sets, with only 71 test examples).

\section{Concluding Remarks}
\label{sec:finalremarks}

In this paper, we presented \modelname\ PT-*, a state-of-the-art foundation model for Portuguese with 900 million parameters, of the encoder class, available in two versions, one for the European Portuguese variant from Portugal (\ptpt) and one for the American Portuguese variant from Brazil (\ptbr). 
To the best of our knowledge, there is no pre-existing encoder specifically developed for \ptpt\ that has been made publicly available and distributed for reuse. Hence, our \modelname\ \ptpt\ is a contribution in that direction and thus sets the state of the art for this variant of Portuguese.
As for \ptbr, our \modelname\ encoder improves the state of the art, taking into account the previous level that was set by the pre-existing encoder \bertimbau, with 330 million parameters, showing superior performance in five out of six downstream tasks used for extrinsic evaluation.

As future work, we will be seeking to progress along a number of directions that may help to secure improvements in the performance of \modelname\ PT-*.
We will experiment with training our encoder versions from scratch on Portuguese data only.
It will be important to keep searching for and using better data in terms of quality (boilerplate cleaning, etc.), coverage of different genres, domains and registers, and coverage of additional Portuguese variants.
And last but not least, we will be trying to obtain better encoders for Portuguese by virtue of improved design, including by increasing their size, experimenting with more architectures, or by finding better hyper-parameters.

\section*{Acknowledgments}

The research reported here was partially supported by: PORTULAN CLARIN—Research Infrastructure for the Science and Technology of Language,
funded by Lisboa 2020, Alentejo 2020 and FCT—Fundação para a Ciência e Tecnologia under the
grant PINFRA/22117/2016; research project ALBERTINA - Foundation Encoder Model for Portuguese and AI, funded by FCT—Fundação para a Ciência e Tecnologia under the
grant CPCA-IAC/AV/478394/2022; innovation project ACCELERAT.AI - Multilingual Intelligent Contact Centers, funded by IAPMEI, I.P. - Agência para a Competitividade e Inovação under the grant C625734525-00462629, of Plano de Recuperação e Resiliência, call RE-C05-i01.01 – Agendas/Alianças Mobilizadoras para a Reindustrialização; and LIACC - Laboratory for AI and Computer Science, funded by FCT—Fundação para a Ciência e Tecnologia under the grant FCT/UID/CEC/0027/2020.

\bibliographystyle{apalike}  
\bibliography{mybibliography}

\end{document}